\documentclass{ifacconf}

\usepackage{graphicx}      
\usepackage{natbib}        

\usepackage{xcolor}
\usepackage{enumitem}
\usepackage{float}
\usepackage{subcaption}
\usepackage{algorithm}
\usepackage{algpseudocode}
\usepackage{booktabs}
\usepackage{lipsum}  
\usepackage{amsfonts}
\usepackage{amsmath}
\usepackage{kotex}

\setlist[enumerate]{leftmargin=*} 

\begin{document}
\begin{frontmatter}

\title{Human-in-the-Loop Neuro-Symbolic Drift Anticipation for Reliable Visual SLAM\thanksref{footnoteinfo}} 

\thanks[footnoteinfo]{This research was supported by Incheon National University Research Grant in 2025 (\#2025-0391).} 

\author{Junhyun Nam$^1$, Wonse Jo$^1$}

\address{$^1$Department of Information and Telecommunication Engineering,}
\address{Incheon National University, Incheon, South Korea 22012. \\
\{namjunhyun,jow\}@inu.ac.kr}

\begin{abstract}                
This paper introduces Hybrid DeepSEE (HDS), a Human-in-the-Loop (HITL) neuro-symbolic framework for proactive drift anticipation in Visual SLAM (V-SLAM). While data-driven models offer predictive power, their ``black-box'' nature often yields physically inconsistent outputs in out-of-distribution (OOD) environments. To address this, HDS integrates neural drift risk estimation with symbolic constraint reasoning. By utilizing a Large Language Model (LLM) as a reasoning bridge, the framework translates qualitative human context into interpretable symbolic constraints. Building upon this architecture, we propose a superior drift anticipation framework that ensures enhanced reliability and consistency in Visual SLAM
\end{abstract}

\end{frontmatter}

\section{Introduction} \label{sec:intro}\vspace{-5pt}
Reliable state estimation is a fundamental requirement for the safe deployment of autonomous mobile robots in mission-critical domains such as industrial automation and defense operations. In such systems, localization accuracy directly affects feedback control performance, trajectory tracking stability, and overall operational safety \citep{5423178}. Visual Simultaneous Localization and Mapping (V-SLAM) has become a widely adopted solution due to its ability to extract rich environmental information using low-cost cameras \citep{7747236}.

Real-world V-SLAM systems, despite their advantages, experience cumulative drift, which refers to the accumulated errors in pose estimation while driving \citep{7747236}. As a result, there is a divergence between the estimated and true trajectories. 
From a control perspective, the estimation degradation can propagate through the feedback loop, leading to performance deterioration or potential instability \citep{RADISAVLJEVICGAJIC2024107015}. Drift is particularly pronounced in challenging environments characterized by repetitive textures, abrupt illumination changes, and low-feature regions, where measurement reliability is significantly reduced \citep{9860754}. 

In safety-critical domains, such as nuclear maintenance, defense operations, and human-robot collaboration, estimation failures can lead to catastrophic consequences including human casualties or severe environmental damage. In these high-stakes contexts, even minor SLAM drift transcends simple positioning errors, potentially triggering abrupt trajectory deviations that propagate into the control loop \citep{7747236}. Therefore, it is essential not only to correct errors after they occur but to anticipate degradation before significant drift compromises the system. Early awareness of drift risk enables proactive supervisory actions---such as motion regulation or re-localization triggers---thereby preserving closed-loop reliability and operational integrity without relying solely on post-hoc, reactive correction \citep{ZHANG2026105306}.

To address this need, we propose a conceptual Neuro-Symbolic Drift Anticipation Framework that introduces a constraint-aware supervisory layer above a conventional V-SLAM estimator using Human-in-the-Loop approach. It monitors drift risk and evaluates consistency under predefined constraints without altering the estimated pose. This supervisory evaluation aims to prevent physically implausible or high-risk estimation conditions.

The main contributions of this work are summarized as follows:
\begin{itemize}
    \item Designing a proactive supervisory framework for anticipating drift risk in V-SLAM-based state estimation.
    \item Proposing a constraint-aware Human-in-the-Loop regulation mechanism that encodes expert geometric and kinematic knowledge through a language-mediated interface, enabling adaptive drift-risk control under explicitly defined safety constraints.
\end{itemize}


\begin{figure}[t]
    \centering

    \includegraphics[width=0.8\columnwidth]{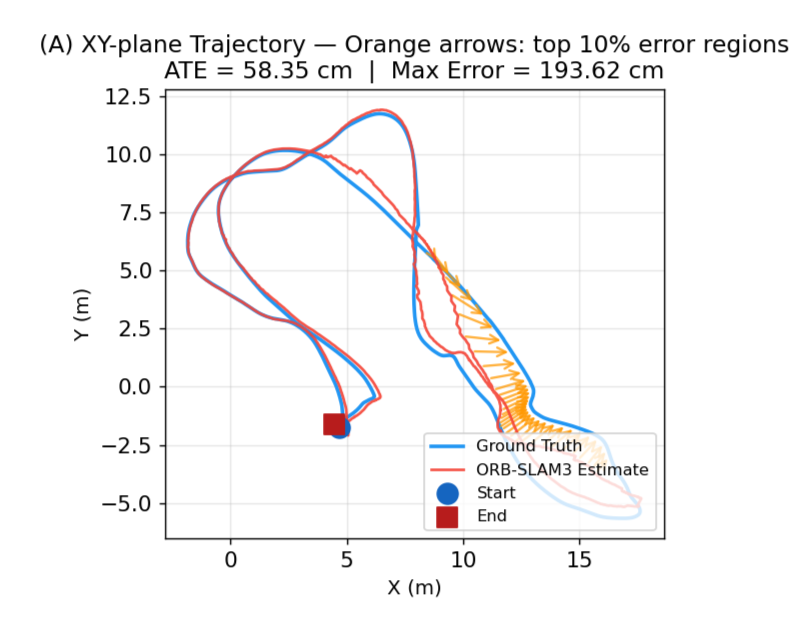} 
    
    \vspace{-10pt} 
    \small\caption{Trajectory comparison between the ground truth and V-SLAM estimation on the EuRoC MH\_04 sequence, illustrating the occurrence of cumulative tracking drift.}
    \label{fig:overall_analysis}
\end{figure}

\section{Related Work}\label{sec:background}
\subsection{Geometric Visual SLAM}
\vspace{-10pt}
Classical visual SLAM systems rely on geometric consistency, feature matching, and nonlinear optimization to estimate camera trajectories. Representative systems such as ORB-SLAM3 \citep{9440682} utilized feature-based tracking, local bundle adjustment, and loop closure detection to refine pose estimates. These approaches achieve high accuracy in structured and well-textured environments through principled optimization over pose graphs. 

However, geometric SLAM systems are fundamentally reactive: accumulated drift is corrected only after sufficient loop evidence is detected. In large-scale exploration or non-revisitable trajectories, loop closure cannot be guaranteed, leading to persistent divergence. Furthermore, feature-based methods degrade significantly in challenging environments—such as low-texture regions or areas with illumination volatility—where reliable feature correspondences become sparse or ambiguous.

Conventional mitigation strategies primarily rely on loop closure detection, which compensates accumulated errors after revisiting previously mapped locations \citep{HO2006740}. Although effective in structured environments, loop closure mechanisms are inherently reactive and provide no guarantee of bounded estimation performance during large-scale exploration or high-speed maneuvers \citep{7747236}. In safety-critical systems, it is insufficient to only rely on the traditional loop closure mechanisms due to safety.

\subsection{Learning-Based Drift and Error Estimation}
\vspace{-10pt}
Recent advances have incorporated deep learning into SLAM pipelines to improve robustness and error awareness. In particular, SEESys--the state-of-the-art online extension of the DeepSEE\citep{10.1145/3570361.3614082} framework--introduced an edge-assisted architecture that achieves real-time pose error estimation with an average end-to-end latency of 37.3 ms \citep{10.1145/3666025.3699341}. By leveraging an attention-based multi-modality model (TimeSformer), SEESys demonstrates the feasibility of delivering instantaneous quality-of-service (QoS) metrics for mobile robots and AR devices.

However, despite their real-time capabilities, recent approaches that integrate deep neural networks (DNNs) into V-SLAM pipelines to enhance perceptual robustness remain fundamentally data-driven black-box systems. While these methods can produce numerically plausible motion estimates, they lack mechanisms to formally enforce geometric or kinematic consistency. As a result, physically inadmissible trajectories may emerge, particularly in out-of-distribution (OOD) environments, where such inconsistencies can arise unpredictably and introduce additional risk into downstream control processes \citep{RADISAVLJEVICGAJIC2024107015}. Furthermore, although systems such as SEESys provide auditory feedback to human operators, they do not offer explainable reasoning nor enable dynamic injection of contextual constraints into the inference loop.

\subsection{Neuro-Symbolic and Constraint-Aware Learning}
\vspace{-10pt}
Neuro-symbolic learning aims to integrate data-driven perception with symbolic reasoning to overcome the limitations of black-box models. Logical Neural Networks (LNNs) enable differentiable reasoning under explicitly defined logical rules, offering improved interpretability and consistency. However, their application to proactive drift anticipation in robotic localization remains largely unexplored.

In contrast to prior work, our framework integrates neural drift risk estimation with symbolic constraint enforcement under a Human-in-the-Loop paradigm. By leveraging a Large Language Model (LLM) as a reasoning bridge to encode expert-defined constraints, we enable a Hybrid DeepSEE (HDS) approach that aligns learned representations with formal safety and kinematic principles.


\section{Design of Hybrid DeepSEE (HDS) Framework} \label{sec:method}

\begin{figure}[t] 
    \centering
    
    \includegraphics[width=1\columnwidth]{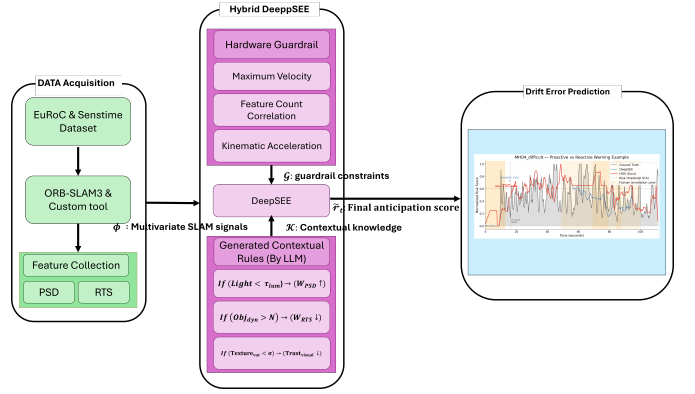}
    
    \vspace{-4pt} 
    \small
    \caption{Overall architecture of the Hybrid DeepSEE (HDS) framework. 
    The system integrates real-time SLAM features with language-mediated symbolic constraints 
    to proactively anticipate drift risk.}
    \label{fig:hds_architecture}
\end{figure}

\noindent 

The proposed Hybrid DeepSEE (HDS) framework proactively anticipates V-SLAM drift by integrating neural perception with human-defined geometric logic. 
As illustrated in Figure~\ref{fig:hds_architecture}, the architecture consists of 
three main modules: 
(i) Neural Drift Risk Estimation, 
(ii) Language-Mediated Symbolic Encoding, and 
(iii) Neuro-Symbolic Supervisory Integration.

\subsection{Problem Formulation}
Let $\hat{P}_t \in SE(3)$ denote the estimated camera pose and $P_t^{*}$ the unknown ground-truth pose. 
We define the integrated drift risk $r_t$ as the expected drift magnitude conditioned on multivariate SLAM signals $\Phi_t$, contextual knowledge $\mathcal{K}$, and guardrail constraints $\mathcal{G}$:

\begin{equation}
r_t = \mathbb{E}[D_t \mid \Phi_t, \mathcal{K}, \mathcal{G}]
\end{equation}

where $\Phi_t$ denotes the extracted SLAM feature set including PSD and RTS, $\mathcal{K}$ represents contextual knowledge derived from language input, and $\mathcal{G}$ denotes hardware-based safety constraints.

\subsection{Neural Drift Risk Estimation}
The SLAM feature set $\Phi_{t}$ comprises high-dimensional internal metrics and environmental descriptors reflecting V-SLAM reliability. Unlike raw geometric features, $\Phi_{t}$ specifically incorporates the following indicators to anticipate estimation degradation:
\begin{itemize}
\setlength\itemsep{0em} 
\item \textbf{Power Spectral Density (PSD):} This metric evaluates signal power distribution across the frequency domain. A decay in high-frequency components indicates texture deprivation or motion blur, serving as a critical precursor to tracking failure and cumulative drift.
\item \textbf{Real-Time Scalability (RTS):} RTS monitors the pipeline's computational health, including tracking latency and hardware utilization. A decrease in RTS signifies a failure to maintain the frame rate necessary for consistent estimation, introducing temporal gaps in the pose graph and leading to trajectory divergence.
\end{itemize}
\vspace{-5pt}

The neural module produces a baseline risk estimate through a deep neural network $f_{\theta}$:
\begin{equation}
\hat{r}_t = f_{\theta}(\Phi_t)
\end{equation}

This value $\hat{r}_t$ reflects data-driven uncertainty based on historical error patterns but, as a purely neural output, it lacks explicit awareness of physical constraints or human-defined safety boundaries.

\subsection{Language-Mediated Symbolic Encoding}
Speech-to-Text (STT) converts human spoken input into textual representations, which are then processed as contextual prompts within the system. 
The Dynamic Logical Updater, powered by LLM, translates these prompts into symbolic rules $\mathcal{R}_k$. For example:

\begin{equation}
\text{Prompt} 
\xrightarrow{\text{LLM}} 
\mathcal{R}_{\text{light}} :
(I_{\text{avg}} < \tau_{\text{lm}}) \Rightarrow (W_{\text{PSD}} \uparrow)
\end{equation}

This pipeline enables adaptive sensitivity adjustment to the environmental factors that may not be fully captured by the neural model.

\subsection{Neuro-Symbolic Supervisory Integration}

The Logical Neural Network fuses neural risk, contextual rules, and guardrail constraints into a final anticipation score $\tilde{r}_{t}$:
  $$\tilde{r}_{t} = \sigma \left( w_{n}\hat{r}_{t} + \sum_{k}\lambda_{k}\mathcal{R}_{k}(K) + \sum_{j}\gamma_{j}\mathcal{G}_{j} \right)$$where $w_{n}$, $\lambda_{k}$, and $\gamma_{j}$ denote weighting coefficients for neural risk scores, symbolic rules, and hardware constraints, respectively. In this conceptual framework, these parameters are systematically initialized to reflect the baseline confidence of neural predictions, the activation frequency of symbolic rules, and the threshold sensitivity of hardware limitations. To preserve the deterministic safety of the system, the dynamic formulation and online adaptive update strategies for these scaling coefficients are designated as crucial directions for future investigation.  The hardware limitations (e.g., maximum velocity, kinematic acceleration limits) act as physical priors that reinforce the risk score when safety thresholds are approached or violated. Importantly, the supervisory integration operates exclusively at the decision-support level and does not modify the pose estimation variables within the SLAM optimization backend. By establishing this neuro-symbolic alignment, the proposed framework provides a structured supervision mechanism that enhances robustness in challenging and Out-of-Distribution (OOD) environments, where purely data-driven or geometric methods often fail unpredictably.

\subsection{Supervisory Drift Monitoring and Decision Support}
The final anticipation score $\tilde{r}_t$ is provided to a 
supervisory monitoring layer that operates independently of 
the underlying SLAM estimator. Importantly, the proposed 
framework does not modify or correct the estimated pose 
$\hat{P}_t$; instead, it evaluates localization reliability 
in real time and provides decision-level guidance. When the 
anticipated drift risk exceeds a predefined threshold, the 
supervisory layer may trigger high-level safety actions such as:
\begin{itemize}
    \item Motion scaling or velocity reduction,
    \item Re-localization request,
    \item Human operator notification.
\end{itemize}
It is important to note that the LLM's reasoning process is 
executed asynchronously only when significant environmental 
changes are detected or explicit user inputs are provided, 
ensuring that high-level symbolic reasoning does not compromise 
the real-time constraints of the underlying V-SLAM backend.

\begin{figure}[t]
    \centering
    
    \includegraphics[width=1\columnwidth]{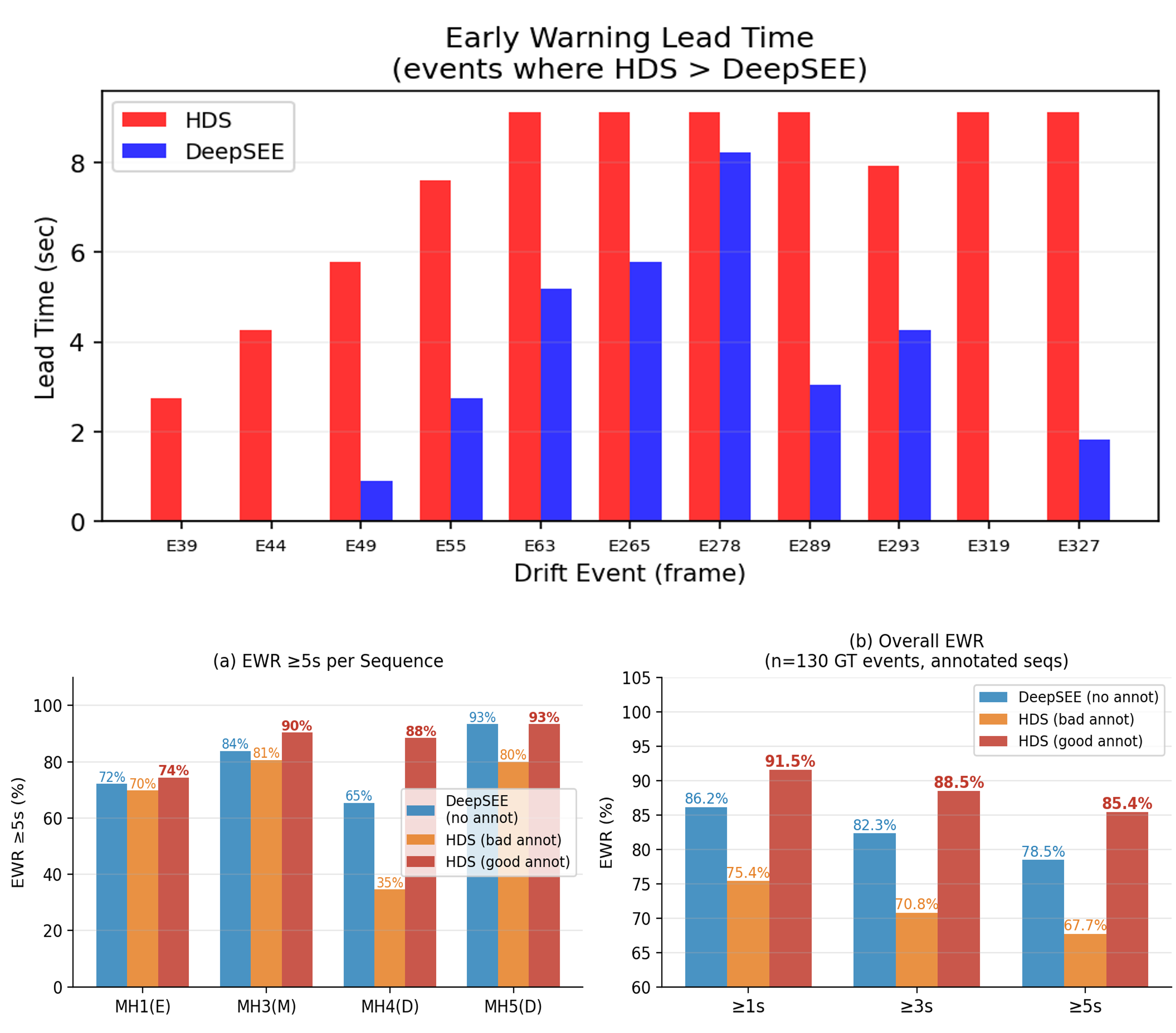} 
    
    \vspace{-10pt} 
    \small
    \caption{Evaluation of the Early Warning Lead Time and Early Warning Rate (EWR) under varying qualities of language-mediated human annotations.}
    \label{fig:ewr_results}
\end{figure}

\noindent

\subsection{Quantitative Evaluation and Experimental Results}

To validate the performance of the proposed Hybrid DeepSEE (HDS) framework, we conducted comparative experiments using the EuRoC MAV dataset, specifically the challenging \textit{MH\_04\_difficult} sequence. We evaluated the proactive drift anticipation capability of HDS against both a standard Random Forest baseline and DeepSEE \citep{10.1145/3570361.3614082}.

The experimental results indicate that HDS generally provides longer Early Warning Lead Times compared to conventional reactive approaches. Across major drift events (e.g., E39 to E327), HDS predicted significant tracking degradation an average of 6.2 seconds in advance, whereas DeepSEE achieved an average lead time of 2.4 seconds or failed to anticipate certain drift events entirely.

Furthermore, we evaluated the Early Warning Rate (EWR) under varying qualities of language-mediated human annotations. Under robust expert-provided contextual annotations (``good annot''), HDS achieved a peak EWR of 91.5\% for a $\geq$1s lead-time window and maintained a relatively high reliability of 85.4\% even under the stricter $\geq$5s condition. In contrast, DeepSEE without contextual assistance showed a noticeable performance decline, dropping to 78.5\% at the $\geq$5s threshold.

However, the experiments also revealed that the overall performance of HDS is highly dependent on the quality and consistency of human-provided language inputs. Accurate and context-rich annotations from expert users significantly improved prediction reliability and anticipation performance, whereas ambiguous or low-quality annotations could reduce system effectiveness. This observation suggests that the proposed framework strongly relies on Human-in-the-Loop interaction, where the quality of linguistic context directly influences the final safety prediction outcome.

These results demonstrate that language-mediated symbolic constraints can help compensate for neural uncertainty and support more reliable and proactive safety regulation in out-of-distribution (OOD) scenarios.

\section{Conclusion}
This paper presents Hybrid DeepSEE (HDS), a human-in-the-loop neuro-symbolic framework designed to proactively anticipate drift errors in Visual SLAM. By fusing data-driven neural perception with language-mediated symbolic reasoning, the architecture effectively addresses the "black-box" limitations and physical inconsistencies prevalent in conventional deep learning models. 

The primary structural contribution of this work lies in decoupling drift-risk anticipation from the underlying state-estimation process, establishing a robust supervisory safety layer that preserves the integrity of the SLAM backend. Crucially, empirical evaluations demonstrate that integrating expert-defined geometric and kinematic constraints via a language interface significantly extends the early warning lead time and enhances prediction reliability. This framework shifts drift monitoring from a reactive recovery process to a proactive, explainable, and trustworthy component of autonomous robotic navigation.

\section{Future Works}

This research presents a conceptual framework, so we outline the following future research plans: \textbf{(1) Adaptive Weight Optimization Strategy:} We will develop an algorithm to determine the optimal weighting coefficients 
    ($w_n, \lambda_k, \gamma_j$) for fusing neural risk, symbolic rules, and hardware constraints. 
    This includes applying adaptive scheduling techniques to increase guardrail weights once the robot approaches physical safety limits, ensuring deterministic safety in high-risk scenarios. 
     \textbf{(2) Validation with physical humanoid:} 
    The proposed HDS framework will be deployed onto a humanoid for real-world testing. We will evaluate the framework's ability to proactively anticipate drift and trigger supervisory interventions within the real-time constraints of the hardware.

\begingroup
\bibliography{ifacconf}
\endgroup

\end{document}